\documentclass{article}
\usepackage{ijcai21}

\usepackage{times}

\usepackage{soul}
\usepackage{url}
\usepackage[hidelinks]{hyperref}
\usepackage[utf8]{inputenc}
\usepackage[small]{caption}
\usepackage{graphicx}
\usepackage{amsmath}
\usepackage{booktabs}
\usepackage{times}
\usepackage{algorithm}
\usepackage{algorithmic}
\usepackage{enumitem}
\usepackage{amsfonts}
\usepackage{dsfont}
\usepackage{amssymb}
\usepackage{amsfonts}
\usepackage{bm}
\usepackage{xspace}
\usepackage{helvet}
\usepackage{linguex}
\usepackage{xcolor}
\usepackage{amsthm}
\usepackage{booktabs}
\usepackage{multirow}
\usepackage[utf8]{inputenc}
\usepackage{color}
\usepackage{diagbox}
\newtheorem{definition}{Definition}
\newtheorem{property}{Property}
\usepackage[justification=centering]{caption}

\title{TE-ESN: Time Encoding Echo State Network for Prediction \\Based on Irregularly Sampled Time Series Data}

\author{
Chenxi Sun$^{1,2}$\and
Shenda Hong$^{3,4}$\and
Moxian Song$^{1,2}$\and\\
Yanxiu Zhou$^{1,2}$\and
Yongyue Sun$^{1,2}$\and
Derun Cai$^{1,2}$\and
Hongyan Li$^{1,2}$\footnote{Contact Author. Peking University, No. 5 Yiheyuan Road, Beijing 100871, People’s Republic of China. (leehy@pku.edu.cn).}\\
\affiliations

$^1$\textit{Key Laboratory of Machine Perception (Ministry of Education), Peking University,} Beijing, China.\\
$^2$\textit{School of Electronics Engineering and Computer Science, Peking University,} Beijing, China.\\
$^3$\textit{National Institute of Health Data Science, Peking University,} Beijing, China.\\
$^4$\textit{Institute of Medical Technology, Health Science Center of Peking University,} Beijing, China.}

\begin{document}

\maketitle

\begin{abstract}
Prediction based on Irregularly Sampled Time Series (ISTS) is of wide concern in the real-world applications. For more accurate prediction, the methods had better grasp more data characteristics. Different from ordinary time series, ISTS is characterised with irregular time intervals of intra-series and different sampling rates of inter-series. However, existing methods have suboptimal predictions due to artificially introducing new dependencies in a time series and biasedly learning relations among time series when modeling these two characteristics. In this work, we propose a novel Time Encoding (TE) mechanism. TE can embed the time information as time vectors in the complex domain. It has the the properties of absolute distance and relative distance under different sampling rates, which helps to represent both two irregularities of ISTS. Meanwhile, we create a new model structure named Time Encoding Echo State Network (TE-ESN). It is the first ESNs-based model that can process ISTS data. Besides, TE-ESN can incorporate long short-term memories and series fusion to grasp horizontal and vertical relations. Experiments on one chaos system and three real-world datasets show that TE-ESN performs better than all baselines and has better reservoir property.
\end{abstract}

\section{Introduction} \label{sec:Introduction}

Prediction based on Time Series (TS) widely exists in many scenarios, such as healthcare management and meteorological forecast \cite{DBLP:journals/sigkdd/XingPK10}. Many methods, especially Recurrent Neural Networks (RNNs), have achieved state-of-the-art \cite{DBLP:journals/datamine/FawazFWIM19}. However, in the real-world applications, TS usually is Irregularly Sampled Time Series (ISTS) data. For example, the blood sample of a patient during hospitalization is not collected at a fixed time of day or week. 
This characteristic limits the performances of the most methods. 

\begin{figure*}[t]
\centerline{\includegraphics[width=0.9\linewidth]{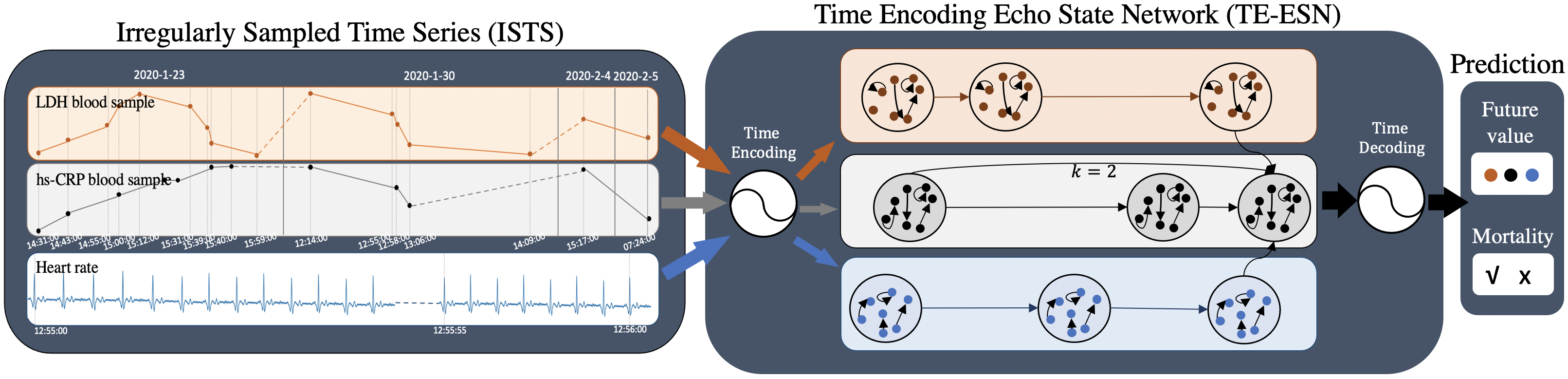}}
\caption{An ISTS example of a COVID-19 patient and the structure of TE-ESN}
\label{fig:main_fig}
\end{figure*}

Basically, a comprehensive learning of the characteristics of ISTS contributes to the accuracy of final prediction \cite{DBLP:conf/ijcai/HaoC20}. For example, the state of a patient is related to a variety of vital signs. ISTS has two characteristics of irregularity under the aspects of intra-series and inter-series:

\begin{itemize}[leftmargin=10 pt]
\setlength{\itemsep}{0pt}
\setlength{\parskip}{0pt}
    \item Intra-series irregularity is the irregular time intervals between observations within a time series. For example, due to the change of patient's health status, the relevant measurement requirements are also changing. For example, in Figure \ref{fig:main_fig}, the intervals between blood sample collections of a COVID-19 patient could be 1 hour or even 7 days. Uneven intervals will change the dynamic dependency between observations and large time intervals will add a time sparsity factor \cite{MRNN}.
    \item Inter-series irregularity is the different sampling rates among time series. For example, in Figure \ref{fig:main_fig}, because vital signs have different rhythms and sensors have different sampling time, for a COVID-19 patient, heart rate is measured in seconds, while blood sample is collected in days. The difference of sampling rates is not conducive to data preprocessing and model design \cite{DBLP:journals/access/KarimMD19}.
\end{itemize}

However, grasping both two irregularities of ISTS is challenging. In the real-world applications, a model usually has multiple time series as input. If seeing the input as a multivariate time series, the data alignment with up/down-sampling and imputation occurs. But it will artificially introduce some new dependencies while omit some original dependencies, causing suboptimal prediction \cite{DBLP:journals/corr/abs-2010-12493}; If seeing the input as multiple separated time series and changing dependencies based on time intervals, the method will encounter the problem of bias, embedding stronger short-term dependency in high sampled time series due to smaller time intervals. This is not necessarily the case, however, for example, although the detection of blood pressure is not frequent than heart rate in clinical practice, its values have strong diurnal correlation \cite{Virk2006Diurnal}.

In order to get rid of the above dilemmas and achieve more accurate prediction, modeling all irregularities of each ISTS without introducing new dependency is feasible. However, the premise is that ISTS can't be interpolated, which makes the alignment impossible, leading to batch gradient descent for multivariate time series hard to implement, aggravating the non-converging and instability of error Back Propagation RNNs (BPRNNs), the basis of existing methods for ISTS \cite{DBLP:journals/corr/abs-2010-12493}. Echo State Networks (ESNs) is a simple type of RNNs and can avoid non-converging and computationally expensive by applying least square problem as the alternative training method \cite{DBLP:conf/nips/Jaeger02}. But ESNs can only process uniform TS by assuming time intervals are equal distributed, with no mechanism to model ISTS. For solving all the difficulties mentioned above, we design a new structure to enable ESNs to handle ISTS data, where a novel mechanism makes up for the disadvantage of no learning of irregularity.

The contributions are concluded as:

\begin{itemize}[leftmargin=10 pt]
\setlength{\itemsep}{0pt}
\setlength{\parskip}{0pt}
    \item We introduce a novel mechanism named Time Encoding (TE) to learn both intra-series irregularity and inter-series irregularity of ISTS. TE represents time points as dense vectors and extends to complex domain for more options of representation. TE injects the absolute and relative distance properties based on time interval and sampling rate into time representations, which helps model two ISTS irregularities at the same time.
    
    \item We design a mode named Time Encoding Encoding Echo State Network (TE-ESN). In addition to the ability of modeling both two ISTS irregularities, TE-ESN can learn the long short-term memories in a time series longitudinally and fuses the relations among time series horizontally.
    
    \item We evaluate TE-ESN for two tasks, early prediction and one-step-ahead forecasting, on MG chaos system, SILSO sunspot dataset, USHCN meteorological dataset and COVID-19 medical dataset. TE-ESN outperforms state-of-the-art models and has better reservoir property.
\end{itemize}

\section{Related Work} \label{sec:related work}

Existing methods for ISTS can be divided into two categories:

The first is based on the perspective of missing data. It discretizes the time axis into non-overlapping intervals, points without data are considered as missing data points. Multi-directional RNN (M-RNN) \cite{MRNN} handled missing data by operating time series forward and backward. Dated Recurrent Units-Decay (GRU-D) \cite{GRUD} used decay rate to weigh the correlation between missing data and other data. However, data imputation may artificially introduce not naturally occurred dependency beyond original relations and totally ignore to model ISTS irregularities.

The second is based on the perspective of raw data. It constructs models which can directly receive ISTS as input. Time-aware Long Short-Term Memory (T-LSTM) \cite{TLSTM} used the elapsed time function for modeling irregular time intervals. Interpolation-Prediction Network \cite{IPN} used three time perspectives for modeling different sampling rates. However, they just performed well in the univariate time series, for multiple time series, they had to apply alignment first, causing the data missing in some time points, back to the defects of the first category.

The adaption of the BPRNNs training requirements causes the above defects. ESNs with a strong theoretical ground, is practical and easy to implement, can avoid non-converging \cite{DBLP:journals/corr/abs-1712-04323,DBLP:journals/corr/abs-2012-02974}. Many state-of-the-art ESNs designs can predict time series well. \cite{2007Optimization} designed a classical reservoir structure using leaky integrator neurons (leaky-ESN) and mitigated noise problem in time series. \cite{Gallicchio2017Deep} 
proposed a stacked reservoirs structure based on deep learning (DL) (DeepESN) to pursue conciseness of ESNs and effectiveness of DL. \cite{DBLP:journals/access/ZhengQLXZM20} proposed a long short-term reservoir structure (LS-ESN) by considering the relations of time series in different time spans. But there is no ESNs-based methods for ISTS.

\section{Time Encoding Echo State Network} \label{sec:Methodology}

The widely used RNN-based methods, especially ESNs, only model the order relation of time series by assuming the time distribution is uniform. We design Time Encoding (TE) mechanism (Section \ref{sec:TE}) to embed the time information and help ESNs to learn the irregularities of ISTS (Section \ref{sec:TE-ESN}). 

\subsection{Definitions} \label{sec:Definitions}

First, we give two new definitions used in this paper.

\begin{definition}[Irregularly Sampled Time Series ISTS] \label{def:ISTS}
A time series $u$ with sampling rate $r_s(d),d\in=\{1,...,D\}$ has several observations distributed with time $t,t\in=\{1,...,T\}$. $u^{d}_{t}$ represents an observation of a time series with sampling rate $r_s(d)$ in time $t$.
\end{definition}

ISTS has two irregularities: (1) Irregular time intervals of intra-series: $t_{i}-t_{i-1}\ne t_{j}-t_{j-1}$. (2) Different sampling rate of inter-series: $r_{s}(d_{i})\ne r_{s}(d_{j})$.

For prediction tasks, one-step-ahead forecasting is using the observed data $u_{1:t}$ to predict the value of $u_{t+1}$, and continues over time; Early prediction is using the observed data $u_{1:t}$ ($t<t_{pre}$) to predict the classes or values in time $t_{pre}$.

\begin{definition}[Time Encoding TE] \label{def:PE}
Time encoding mechanism aims to design methods to embed and represent every time point information of a time line. 
\end{definition}

TE mechanism extends the idea of Positional Encoding (PE) in natural language processing. PE was first introduced to represent word positions in a sentence \cite{DBLP:conf/icml/GehringAGYD17}. Transformer \cite{DBLP:conf/nips/VaswaniSPUJGKP17} model used a set of sinusoidal functions discretized by each relative input position, shown in Equation \ref{eq:PE}. Where $pos$ indicates the position of a word, $d_{model}$ is the embedding dimension. Meanwhile, a recent study \cite{DBLP:conf/iclr/WangZLLZS20} encoded word order in complex embeddings. An indexed $j$ word in the $pos$ position is embeded as $ g_{pe}(j,pos)=re^{i\omega_{j}pos+\theta_{j}}$.  $r$, $\omega$ and $\theta$ denote amplitude, frequency and primary phase respectively. They are all the parameters that should to be learned using deep learning model.

\begin{equation} \label{eq:PE} 
\small
\left\{ \begin{aligned} 
&PE(pos,2i)=\sin (\frac{pos}{10000^{\frac{2i}{d_{model}}}}) \\
&PE(pos,2i+1)=\cos (\frac{pos}{10000^{\frac{2i}{d_{model}}}})
\end{aligned} \right.
\end{equation}

\subsection{Time encoding mechanism} \label{sec:TE}

First, we introduce how the Time Vector (TV) perceives irregular time intervals of a single ISTS with the fixed sampling rate. Then, we show how Time Encoding (TE) embeds time information of multiple ISTS with different sampling rates. The properties and proofs are summarized in the Appendix.

\subsubsection{Time vector with fixed sampling rate} \label{sec:TE_fixed}

Now, let's only consider one time series, whose irregularity is just reflected in the irregular time intervals. Inspired by Positional Encoding (PE) in Equation \ref{eq:PE}, we apply Time Vector (TV) to note the time codes. Thus, in a time series, each time point is tagged with a time vector:

\begin{equation} \label{eq:TV}
\begin{aligned}
\small
TV(t)=[..., \sin &(c_{i}t), \cos (c_{i}t),...]\\
c_{i}=MT^{-\frac{2i}{d_{TV}}}, &i=0,...,\frac{d_{TV}}{2}-1
\end{aligned}
\end{equation}

In Equation \ref{eq:TV}, each time vector has $d_{TV}$ embedding dimensions. Each dimension corresponds to a sinusoidal. Each sinusoidal wave forms a geometric progression from $2\pi$ to $MT\pi$. $MT$ is the biggest wavelength defined by the maximum number of input time points.

Without considering the different sampling rates of inter-series, for a single ISTS, TV can simulate the time intervals between two observations by its properties of absolute distance and relative distance.

\begin{property}[Absolute Distance Property] \label{property:Absolute}
For two time points with distance $p$, the time vector in time point $t+p$ is the linear combination of the time vector in time point $t$. 
\begin{equation} \label{eq:absolute}
\small
\begin{aligned} 
 TV(t+p) =(a,&b)\cdot TV(t) \\
 a=TV(p,2i+1), &b=TV(p,2i)
\end{aligned}
\end{equation}
\end{property}

\begin{property}[Relative Distance Property] \label{property:Relative}
The product of time vectors of two time points $t$ and $t+p$ is negatively correlated with their distance $p$. The larger the interval, the smaller the product, the smaller the correlation.
\begin{equation} \label{eq:relative}
\small
TV(t)\cdot TV(t+p)=\sum_{i=0}^{\frac{d_{TV}}{2}-1}\cos(c_{i}p)
\end{equation}
\end{property}

For a computing model, if its inputs have the time vectors of time points corresponding to each observation, then the calculation of addition and multiplication within the model will take the characteristics of different time intervals into account through the above two properties, improving the recognition of long term and short term dependencies of ISTS. Meanwhile, without imputing new data, natural relation and dependency within ISTS are more likely to be learned.

\subsubsection{Time encoding with different sampling rates} \label{sec:TE_different}

When the input is multi-series, another irregularity of ISTS, different sampling rates, shows up. Using the above introduced time vector will encounter the problem of bias. It will embed more associations between observations with high sampling rate according to the Property \ref{property:Relative}, as they have smaller time intervals. But we can not simply conclude that the correlation between the values of time series with low sampling rate is weak.

Thus, we design an advanced version of time vector, noted Time Encoding (TE), to encode time within multiple ISTS. TE extends TV to complex-valued domain. For a time point $t$ in the $d$-th ISTS with $r_{s}(d)$ sampling rate, the time code is in Equation \ref{eq:TED}, where $\omega$ is the frequency.

\begin{equation} \label{eq:TED}
\small
 TE(d,t)=e^{i(\omega t)}, \omega=\omega_{d}\cdot r^{-1}_{s}(d)
\end{equation}

Compared with TV, TE has two advantages:

The first is that TE not only keeps the property \ref{property:Absolute} and \ref{property:Relative}, but also incorporates the influence of frequency $\omega$, making time codes consistent at different sampling rates.

$\omega$ reflects the sensitivity of observation to time, where a large $\omega$ leads to more frequent changes of time codes and more difference between the representations of adjacent time points. For relative distance property, a large $\omega$ makes the product large when distance $p$ is fixed.

\begin{property}[Relative Distance Property with $\omega$] \label{property:Relative_omega}
The product of time encoding of two time points $t$ and $t+p$ is positive correlated with frequency $\omega$. 
\begin{equation} \label{eq:relative_omega}
\small
    TE(t)\cdot TE(t+p)=e^{i\omega t}\cdot e^{i\omega(t+p)}=e^{i\omega(2t+p)}
\end{equation}
\end{property}

In TE, we set $\omega=\omega_{d}\cdot r^{-1}_s(d)$. $\omega_{d}$ is the frequency parameter of $d$-th sampling rate. TE fuses the sampling rate term $r^{-1}_s(d)$ to avoid the bias of time vector causing by only considering the effect of distance $p$.

The second is that each time point can be embeded into $d_{TE}$ dimensions with more options of frequencies by setting different $\omega_{j,k}$ in Equation \ref{eq:TEV}. 

\begin{equation} \label{eq:TEV}
\small
\begin{aligned}
TE(d,t)=e^{i(\omega t)}, &\omega=\omega_{j,k}\cdot r^{-1}_{s}(d)\\
j=0,...,d_{TV}-1,&k=0,...,K-1
\end{aligned}
\end{equation}

In TE, $\omega_{j,k}$ means the time vector in dimension $j$ has $K$ frequencies. But in Equation \ref{eq:TV} of TV, the frequency of time vector in dimension $i$ is fixed with $c_{i}$.

\subsubsection{The relations between different mechanisms} \label{sec:TE_relation}
Time encoding with different sampling rates is related to time vector with fixed sampling rate and a general complex expression \cite{DBLP:conf/iclr/WangZLLZS20}.

\begin{itemize}[leftmargin=10 pt]
\setlength{\itemsep}{0pt}
\setlength{\parskip}{0pt}
    \item TV is a special case of TE. If we set $\omega_{k,j}=c_{i}$, then $TE(d,t)=TV(t,2i+1)+iTV(t,2i)$. 
    \item TE is a special case of a fundamental complex expression $r\cdot e^{i\cdot (\omega x+\theta)}$. We set $\theta=0$ as we focus more on the relation between different time points than the value of the first point; We understand term $r$ as the representation of observations and leave it to learn by computing models. Besides, TE inherits the properties of position-free offset transformation and boundedness \cite{DBLP:conf/iclr/WangZLLZS20}.
\end{itemize}

\subsection{Injecting time encoding mechanism into echo state network}\label{sec:TE-ESN}

Echo state network is a fast and efficient recurrent neural network. A typical ESN consists of an input layer $W_{in}\in R^{N\times D}$, a recurrent layer, called reservoir $W_{res}\in R^{N\times N}$, and an output layer $W_{out}\in R^{M\times N}$. The connection weights of the input layer and the reservoir layer are fixed after initialization, and the output weights are trainable. $u(t)\in R^{D}$, $x(t)\in R^{N}$ and $y(t)\in R^{M}$ denote the input value, reservoir state and output value at time $t$, respectively. The state transition equation is:

\begin{equation} \label{eq:ESN_trans}
\small
\begin{aligned}
x(t)=f (W_{in}u(t)&+W_{res}x(t-1))\\
y(t)=&W_{out}x(t)   
\end{aligned}
\end{equation}

\begin{algorithm}[t]\caption{TE-ESN} \label{alg:TE-ESN}
\begin{algorithmic}[1] 
\small
\REQUIRE 
$U_{train}=\{u^d_t,t\}$: training input;\\
$Y_{train}={y^d_t,t}$: teacher signal;\\
$U_{test}=\{u^d_t,t\}$: test input;\\
$MT$: maximum time;\\
$\gamma_{l}$: leaky rate; $\gamma_{f}$: fusion rate; \\
$k$: long term time span; \\
$\lambda$: regularization coefficient;\\
$w_{in}$: input scale of $W_{in}$;\\
$\rho(W_{res})$: spectral radius of $W_{res}$;\\
$\alpha$: sparsity of $W_{res}$;\\

\ENSURE 
$Y_{pre}$: prediction result.

\STATE Randomly initialized $W_{in}$ in $[-w^{in},w^{in}]$;
\STATE Randomly initialized $W_{res}$ with $\alpha$ and $\rho(W_{res})$.

\FOR {$i = 1$ to $|U_{train}|$}
\FOR {$t = 1$ to $MT$}
\STATE Compute $TE(d,t)$ by Equation \ref{eq:TEV}
\STATE Compute $\tilde{x}(t)$ by Equation \ref{eq:TE-ESN_xtrans}
\ENDFOR
\ENDFOR
\STATE $\tilde{X}=\{\tilde{x}(t)\}$
\STATE $TE=\{TE(t)\}$
\STATE Compute $W_{out}$ by Equation \ref{eq:TE-ESN_objective}

\FOR {$t = 1$ to $T_{test}$}
\STATE Compute $TE_{test}(d,t)$ by Equation \ref{eq:TEV}
\STATE Compute $\tilde{x}_{test}(t)$ by Equation \ref{eq:TE-ESN_xtrans}
\ENDFOR
\STATE $\tilde{X}_{test}=\{\tilde{x}_{test}(t)\}$
\STATE $TE_{test}=\{TE_{test}(t)\}$
\STATE $Y_{pre}=W_{out}(\tilde{X}_{test}-TE_{test})$
\end{algorithmic}\nonumber
\end{algorithm} 

Before training, three are three main hyper-parameters of ESNs: Input scale $w^{in}$; Sparsity of reservoir weight $\alpha$; Spectral radius of reservoir weight $\rho(W_{res})$ \cite{DBLP:journals/corr/abs-1910-04426}.

However, existing ESNs-based methods cannot model the irregularities of ISTS. Thus, we make up for this by proposing Time Encoding Echo State Network (TE-ESN).

\textbf{Time Encoding.} TE-ESN has $D$ reservoirs, assigning each time series of input an independent reservoir. An observation $u^{d}_t$ is transferred trough input weight $W^{d}_{in}$, time encoding $TE(d,t)$, reservoir weight $W^{d}_{res}$ and output weight $W^{d}_{out}$. The structure of TE-ESN is shown in Figure \ref{fig:main_fig}. The state transition equation is:

\begin{equation} \label{eq:TE-ESN_xtrans}
\setlength{\abovedisplayskip}{0pt}
\setlength{\belowdisplayskip}{0pt}
\small
\begin{aligned}
\tilde{x}^{d}_t=\gamma_{f} {x^{d}_t}' + (1-\gamma_{f}) x^{D\setminus d} \quad \textit{Reservoir state}\\
{x^{d}_t}' =\gamma_{l} x^{d}_t + (1-\gamma_{l})(x^{d}_{t-1}+  x^{d}_{t-k}) \quad \textit{Long short state}\\
x^{d}_t= \tanh (TE(d,t)+W^{d}_{in}u^{d}_t+W^{d}_{res}\tilde{x}^{d}_{t-1})\quad \textit{Time encoding state}\\
x^{D\setminus d}= \frac{1}{D-1} \sum_{i\in D\setminus d} \tilde{x}^{i}  \quad \textit{Neighbor state}
\end{aligned}
\end{equation}

TE-ESN creates three highlights compared with other ESNs-based methods by changing the \textit{Reservoir state}:

\begin{itemize}[leftmargin=10 pt]
\setlength{\itemsep}{0pt}
\setlength{\parskip}{0pt}
    \item Time encoding mechanism (TE). TE-ESN integrates time information when modeling the dynamic dependencies of input, by changing recurrent states in reservoirs through TE term to \textit{Time encoding state}.
    \item Long short-term memory mechanism (LS). TE-ESN leans different temporal span dependencies, by incorporating not only short-term memories from state in last time, but also long-term memories from state in former $k$ time ($k$ is the time skip) to \textit{Long short state}.
    \item Series fusion (SF). TE-ESN also considers the horizontal information between time series, by changing \textit{Reservoir state} according to not only the former state in its time series but also the \textit{Neighbor state} in other time series.
\end{itemize}

The coefficients $\gamma_{l}$ and $\gamma_{f}$ trade off the memory length in \textit{Long short state} and the fusion intensity in \textit{Reservoir state}. 

\textbf{Time Decoding.} The states in reservoir of TE-ESN have time information as TE embeds time codes into the representations of model input. For final value prediction, it should decode the time information and get the real estimated value at time $t_{pre}$ by Equation \ref{eq:TE-ESN_ytrans}. Further, by changing the time $t_{pre}$, we can get different prediction results in different time points.

\begin{equation} \label{eq:TE-ESN_ytrans}
\small
y(t_{pre}) = W_{out}(\tilde{x}(t)-TE(t_{pre}))
\end{equation}

Equation \ref{eq:TE-ESN_objective} is the calculation formula of the readout weights when training to find a solution to the least squares problem with regularization parameter $\lambda$.

\begin{equation} \label{eq:TE-ESN_objective}
\small
\begin{aligned}
\min_{W_{out}}||Y_{pre}&-Y||^2_2+\lambda ||W_{out}||^{2}_{2}\\
W_{out} = Y(\tilde{X} - TE)^{T}((\tilde{X}& - TE)(\tilde{X}- TE)^{T}+\lambda I)^{-1}
\end{aligned}
\end{equation}

Algorithm \ref{alg:TE-ESN} shows the process of using TE-ESN for prediction. Line 1-11 obtains the solution of readout weights $W_{out}$ of TE-ESN by that using the training data. Line 12-18 shows the way to predict the value of test data. Assuming the reservoir size of TE-ESN is fixed by $N$, The maximum time $MT$ is $T$, the input has $D$ time series, the complexity is:

\begin{equation} \label{eq:TE-ESN_complexity}
\small
C=O(\alpha TN^{2}+TND)
\end{equation}

\begin{figure*}[t]
\centerline{\includegraphics[width=0.95\linewidth]{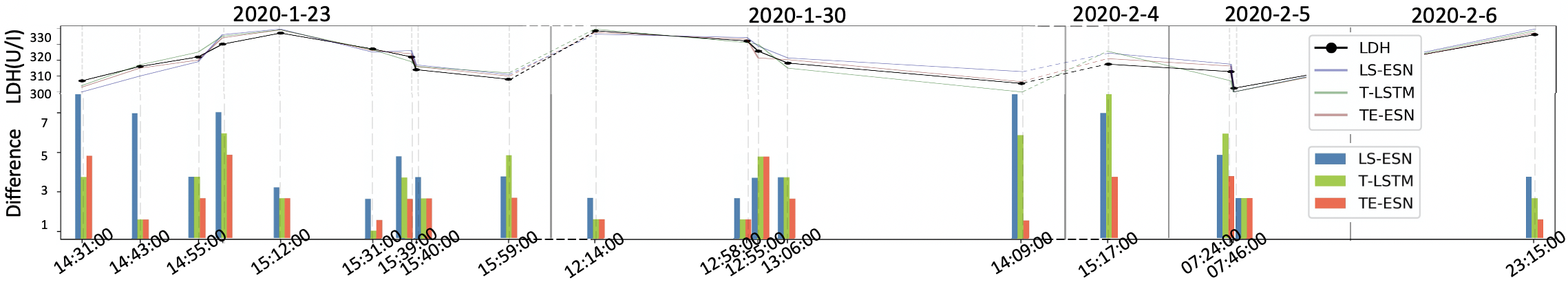}}
\caption{Lactic dehydrogenase (LDH) forecasting for a 70-year-old female COVID-19 patient}
\label{fig:results_case}
\end{figure*}

\begin{table*}[t]
\caption{Prediction results of nine methods on four datasets (COVID-19 mortality in AUC-ROC; Others in MSE)}\label{tb:results}
\scriptsize
\centering
\begin{tabular}{llllllllll}
\hline
& &BPRNNs-based & & & &ESNs-based & &\\
\hline
&M-RNN &T-LSTM &GRU-D &ESN &leaky-ESN &DeepESN &LS-ESN &TV-ESN &\textbf{TE-ESN}\\
\hline
MGsystem   &0.232±0.005 &0.216±0.0.003 &0.223±0.005 &0.229±0.001 &0.213±0.0.001 &\underline{0.197±0} &0.198±0 &0.204±0.001 &\textbf{0.195±0.001}\\
\hline 
SILSO ($10^{-6}$)  &2.95±0.74 &2.93±0.81 &2.99±0.69 &3.07±0.63 &2.95±0.59   &2.80±0.73 &\underline{2.54±0.69} &\underline{2.54±0.79} &\textbf{2.39±0.78}\\
\hline
USHCN   &0.752±0.32 &0.746±0.33 &0.747±0.25 &0.868±0.29  &0.857±0.20 &\underline{0.643±0.12} &0.663±0.15 &0.647±0.15 &\textbf{0.640±0.19} \\
\hline
\multirow{2}{*}{COVID-19}  &0.959±0.004 &\underline{0.963±0.003} &0.963±0.004 &0.941±0.003 &0.942±0.003 &0.948±0.003 &0.949±0.003 &0.958±0.002 &\textbf{0.965±0.002}\\
&0.098±0.0.005 &\underline{0.096±0.007} &0.100±0.005 &0.136±0.006 &0.135±0.007 &0.129±0.006 &0.120±0.007 &0.115±0.005 &\textbf{0.093±0.005}\\
\hline
\end{tabular}
\end{table*}

\section{Experiments}\label{sec:Experiments}

\subsection{Datasets} \label{sec:Datasets}

\begin{itemize}[leftmargin=10 pt]
\setlength{\itemsep}{0pt}
\setlength{\parskip}{0pt}

    \item \textit{MG system} \cite{Mackey1977Oscillation}. Mackey-Glass system is a classical chaotic system, often used to evaluate ESNs. $y(t+1)=y(t)+\delta(a\frac{y(t-\frac{\tau}{\delta})}{1+y(t-\frac{\tau}{\delta})^{n}}-by(t))$. We initialized $\delta,a,b,n,\tau,y(0)$ to $0.1,0.2,-0.1,10,17,1.2$, $t$ random increases with irregular time interval. The task is one-step-ahead-forecasting in first 1000 time.
    
    \item \textit{SILSO} \cite{SILSO}. SILSO provides an open-source monthly sunspot series from 1749 to 2020. It has irregular time intervals, from 1 to 6 month. The task is one-step-ahead forecasting from 1980 to 2019. 
    
    \item \textit{USHCN} \cite{USHCN}. The dataset consists of daily meteorological data of 48 states from 1887 to 2014. We extracted the records of temperature, snowfall and precipitation from New York, Connecticut, New Jersey and Pennsylvania. Each TS has irregular time intervals, from 1 to 7 days. Sampling rates are different among TS, from 0.33 to 1 per day. The task is to early predict the temperature of New York in next 7 days.
    
    \item \textit{COVID-19} \cite{COVID-19}. The COVID-19 patients’ blood samples dataset were collected between 10 Jan. and 18 Feb. 2020 at Tongji Hospital, Wuhan, China, containing 80 features from 485 patients with 6877 records. Each TS has irregular time intervals, from 1 minus to 12 days. Sampling rates are different among TS, from 0 to 6 per day. The task is to early predict in-hospital mortality before 24 hours and one-step-ahead forecasting for each biomarkers. 

\end{itemize}

\subsection{Baselines} \label{sec:Baselines}

The code of 9 baselines with 3 categories is available at \url{https://github.com/PaperCodeAnonymous/TE-ESN}.

\begin{itemize}[leftmargin=10 pt]
\setlength{\itemsep}{0pt}
\setlength{\parskip}{0pt}
    \item \textit{BPRNNs-based}: There are 3 methods designed for ISTS data with BP training - M-RNN \cite{MRNN}, T-LSTM \cite{TLSTM} and GRU-D \cite{GRUD}. Each of them have be introduced in Section \ref{sec:related work}.
    
    \item \textit{ESNs-based}: There are 4 methods designed based on ESNs - ESN \cite{DBLP:conf/nips/Jaeger02}, Leaky-ESN \cite{2007Optimization}, DeepESN \cite{Gallicchio2017Deep} and LS-ESN \cite{DBLP:journals/access/ZhengQLXZM20}. Each of them have be introduced in Section \ref{sec:related work}. 
    
    \item \textit{Our methods}: We use TV-ESN with the time representation embedded by TV, we use TE-ESN with the time representation embedded by TE. 
    
\end{itemize}

\subsection{Experiment setting} \label{sec:Experimentsetting}

We use Genetic Algorithms (GA) \cite{DBLP:journals/ijon/ZhongXLW17} to optimize hyper-parameters shown in Table \ref{tb:hyper-parameters_results}. For TV-ESN, we set $\omega=c_{i},d_{TV}=64$. For TE-ESN, We set $\omega_{k,j}=M_{j}^{-\frac{2j}{d_{TE}}}$, where $M_{0}=\frac{MT}{2},M_{1}=MT,M_{2}=2MT,M_{3}=4MT$ and $d_{TE}=64$. Results are got by 5-fold cross validation. Method performances are evaluated by the Area Under Curve of Receiver Operating Characteristic (AUC-ROC) (higher is better) and the Mean Squared Error (MSE) (lower is better). Network property of ESNsis evaluated by Memory Capability (MC) (higher is better) \cite{DBLP:journals/nn/FarkasBG16} in Equation \ref{eq:MC}. where $r^{2}$ is the squared correlation coefficient. 

\begin{table}[t]
\caption{Search settings of hyper-parameters}\label{tb:Hyper-parameters}
\scriptsize
\centering
\begin{tabular}{lllll}
\hline
&Parameters &Value range &Parameters &Value range\\
\hline
&$w^{in},\alpha,\rho$ &$(0,1]$ &$\gamma_{l},\gamma_{f}$ &$[0,1]$\\
&$k$ &$\{2,4,6,8,10,12\}$ &$\lambda$ &$\{10^{-4},10^{-2},1\}$\\
\hline
\end{tabular}
\end{table}

\begin{equation} \label{eq:MC}
\small
MC=\sum_{k=0}^{\infty}r^{2}(u(t-k),y(t))
\end{equation}

\subsection{Results} \label{sec:Results}

We show the results from five perspectives below. The conclusions drawn from the experimental results are shown in italics. More experiments are in Appendix.

\subsubsection{Prediction results of methods}

Shown in Table \ref{tb:results}: (1) TE-ESN outperforms all baselines on four datasets. It means \textit{Learning two irregularities of ISTS helps for prediction and TE-ESN has this ability}. (2) TE-ESN is better than TV-ESN in multivariable time series datasets (COVID-19, USHCN) shows \textit{the effect of Property \ref{property:Relative_omega} of TE}; TE-ESN is better than TV-ESN in univariable time series datasets (SILSO, MG) shows \textit{the advantage of multiple frequencies options of TE}. (3) ESNs-based methods perform better in USHCN, SILSO and MG, while BPRNNs-based method performs better in COVID-19. Which shows \textit{the characteristic of ESNs that they are good at modeling the consistent dynamic chaos system}, such as astronomical, meteorological and physical. Figure \ref{fig:results_case} shows a case of forecasting lactic dehydrogenase (LDH), an important bio-marker of COVID-19 \cite{COVID-19,DBLP:journals/BMC/sun}. TE-ESN has smallest difference between real and predicted LDH values.

\subsubsection{Time encoding mechanism analysis}

\begin{figure}[!ht]
\centerline{\includegraphics[width=0.93\linewidth]{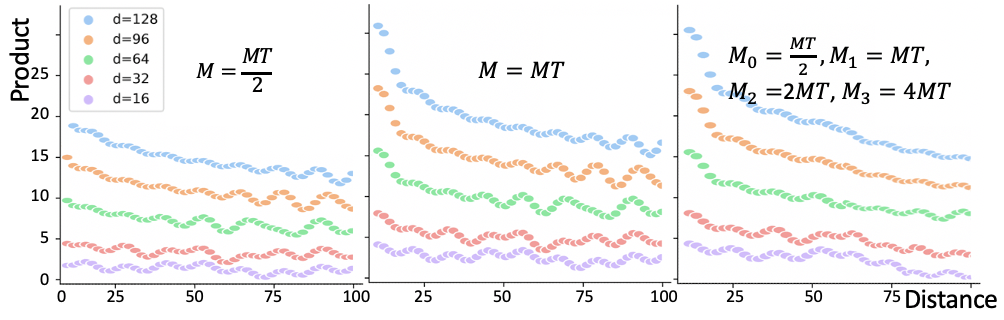}}
\caption{Dimension and frequency setting of time encoding}
\label{fig:TE_dimension}
\end{figure}

Dot product between two sinusoidal positional encoding decreases with increment of absolute value of distance \cite{DBLP:journals/corr/abs-1911-04474}. (1) Figure \ref{fig:TE_dimension} shows the relation of TE dot product and time distance, it shows that \textit{using multiple frequencies will enhance monotonous of negative correlation between dot product and distance}. (2) Table \ref{tb:TEsetting_results} shows the prediction results in different TE settings, results shows that \textit{using multiple frequencies can improve the prediction accuracy}.

\begin{table}[t]
\caption{Prediction results of TE-ESN with different $\omega,d_{TE}$}\label{tb:TEsetting_results}
\scriptsize
\centering
\begin{tabular}{lllll}
\hline
&$c_{i},32$ &$c_{i},64$ &$\omega_{d,i},32$ &$\omega_{d,i},64$ \\
\hline
MGsystem   &0.226±0.0.001 &0.204±0.001 &0.210±0.001 &\textbf{0.193±0.001}\\
\hline 
SILSO    &2.69±0.60 &2.54±0.79 &2.55±0.75 &\textbf{2.39±0.78}\\
\hline
USHCN   &0.681±0.18  &0.670±0.20 &0.673±0.17 &\textbf{0.640±0.19} \\
\hline
\multirow{2}{*}{COVID-19}   &0.949±0.002 &0.952±0.003 &0.950±0.002 &\textbf{0.965±0.002}\\
 &0.105±0.006 &0.099±0.005 &0.101±0.005 &\textbf{0.093±0.005}\\
\hline
\end{tabular}
\end{table}

\subsubsection{Ablation study of TE-ESN}

We test the effect of TE, LS and SF, which are introduced in Section \ref{sec:TE-ESN}, by removing TE term, setting $\gamma_{l}=1$ and setting $\gamma_{f}=1$. The results in Table \ref{tb:ablation_results} show that \textit{all theses three mechanisms of TE-ESN contribute to the final prediction tasks}. TE has the greatest impact in COVID-19, the reason may be that the medical dataset has the strongest irregularity compared with other datasets. LS has the greatest impact in USHCN and SILSO, as there are many long time series, it is necessary to learn the dependence in different time spans. SF has a relatively small impact, the results have no change in SILSO and MG as they are univariate. 

\begin{table}[t]
\caption{Prediction results of TE-ESN with different mechanisms}\label{tb:ablation_results}
\scriptsize
\centering
\begin{tabular}{lllll}
\hline
&w/o TE &w/o LS &w/o SF  &\textbf{TE-ESN}\\
\hline
MGsystem   &0.210±0.0.001 &0.213±0.001 &0.193±0.001 &\textbf{0.193±0.001}\\
\hline 
SILSO    &2.79±0.63 &2.93±0.69 &2.39±0.78 &\textbf{2.39±0.78}\\
\hline
USHCN   &0.713±0.12  &0.757±0.21 &0.693±0.16 &\textbf{0.640±0.19} \\
\hline
\multirow{2}{*}{COVID-19}   &0.943±0.003 &0.949±0.003 &0.956±0.003 &\textbf{0.965±0.002}\\
 &0.135±0.006 &0.130±0.006 &0.125±0.007 &\textbf{0.093±0.005}\\
\hline
\end{tabular}
\end{table}

\subsubsection{Hyper-parameters analysis of TE-ESN}

\begin{figure}[!ht]
\centerline{\includegraphics[width=0.8\linewidth]{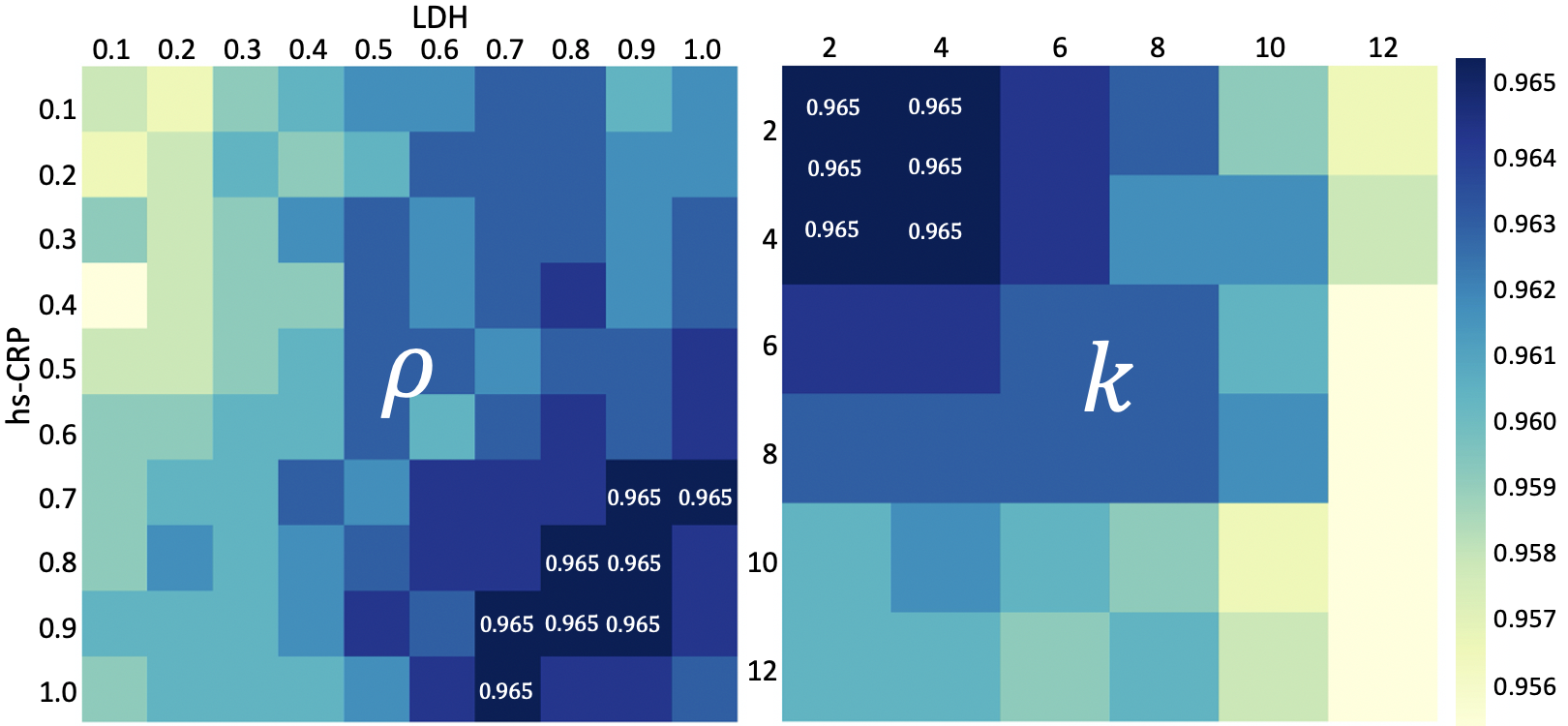}}

\caption{Mortality prediction of TS-ESN with different $\rho$ and $k$}
\label{fig:rho_k}
\end{figure}

In TE-ESN, each time series has a reservoir, reservoirs setting can be different. Figure \ref{fig:rho_k} shows COVID-19 mortality prediction results when changing spectral radius $\rho$ and time skip $k$ of LDH and hs-CRP. \textit{Setting uniform hyper-parameters or different hyper-parameters for each reservoir has little effect on the prediction results}. Thus, we set all reservoirs with the same hyper-parameters for efficiency. Table \ref{tb:hyper-parameters_results} shows the best hyper-parameter settings. 

\begin{table}[t]
\caption{Best settings of hyper-parameters of TE-ESN}\label{tb:hyper-parameters_results}
\scriptsize
\centering
\begin{tabular}{llllllll}
\hline
&$w^{in}$ &$\alpha$ &$\rho$ &$\gamma_{l}$ &$k$  &$\gamma_{f}$ &$\lambda$\\
\hline
MGsystem &1  &0.1 &0.7 &0.8 &6 &1.0 &$10^{-2}$\\
\hline 
SILSO  &1  &0.1 &0.6 &0.8 &10 &1.0 &$10^{-2}$\\
\hline
USHCN &1  &0.1  &0.7 &0.8 &12 &0.8 &$10^{-2}$\\
\hline
\multirow{2}{*}{COVID-19}  &1 &0.3 &0.9 &0.7 &4 &0.9 &$10^{-2}$\\
&1 &0.2 &0.8 &0.8 &2 &0.8 &$10^{-2}$\\
\hline
\end{tabular}
\end{table}

\subsubsection{Memory capability analysis of TE-ESN}
Memory capability (MC) can measure the short-term memory capacity of reservoir, an important property of ESNs \cite{DBLP:journals/nn/GallicchioMP18}. Table \ref{tb:MC} shows that TE-ESN obtains the best MC, and \textit{TE mechanism can increase the memory capability}.

\begin{table}[t]
\caption{Memory capacity results of ESNs-based methods}\label{tb:MC}
\scriptsize
\centering
\begin{tabular}{lllllll}
\hline 
ESN &leaky-ESN & DeepESN &LS-ESN &TE-ESN (-TE) &\textbf{TE-ESN}\\
\hline
35.05 &39.65 &42.98 &46.05 &40.46 &\textbf{47.83}\\
\hline
\end{tabular}
\end{table}

\section{Conclusions}\label{sec:Conclusions}

In this paper, we propose a novel Time Encoding (TE) mechanism in complex domain to model the time information of Irregularly Sampled Time Series (ISTS). It can represent the irregularities of intra-series and inter-series. Meanwhile, we create a novel Time Encoding Echo State Network (TE-ESN), which is the first method to enable ESNs to handle ISTS. Besides, TE-ESN can model both longitudinal long short-term dependencies in time series and horizontal influences among time series. We evaluate the method and give several model related analysis in two prediction tasks on one chaos system and three real-world datasets. The results show that TE-ESN outperforms the existing state-of-the-art models and has good properties. Future works will focus more on the dynamic reservoir properties and hyper-parameters optimization of TE-ESN, and will incorporate deep structures to TE-ESN for better prediction accuracy. 

\bibliographystyle{named}
\bibliography{ijcai21}

\end{document}